\documentclass[10pt,journal,compsoc]{IEEEtran}

\ifCLASSOPTIONcompsoc
  \usepackage[nocompress]{cite}
\else
  \usepackage{cite}
\fi

\ifCLASSINFOpdf

\else

\fi

\usepackage{graphicx}
\usepackage{hyperref}
\usepackage{url}
\usepackage{amsmath}
\usepackage{algorithm,algorithmic}
\hypersetup{
    colorlinks=true,
    linkcolor=blue,
    filecolor=magenta,      
    urlcolor=cyan,
}

\renewcommand{\arraystretch}{1.1}
\usepackage{array} 
\newcolumntype{P}[1]{>{\centering\arraybackslash}p{#1}}
\usepackage{threeparttable,booktabs}

\newcommand{\ie}{\textit{i.e.}}
\usepackage{doi}
\usepackage{epsfig} 
\usepackage{mathrsfs} 
\usepackage{mwe}
\usepackage{graphbox}
\usepackage{xcolor}
\usepackage{colortbl} 
\usepackage{caption}  
\usepackage{subcaption}   
\usepackage{datatool,filecontents}
\newcolumntype{P}[1]{>{\centering\arraybackslash}p{#1}} 
\DTLloaddb[keys={meth,colA,colB,colC,colD,colE,colF,colG,colH}]{fig2}{images/fig2/fig2.csv}
\DTLloaddb[keys={meth,colA,colB,colC,colD,colE,colF}]{fig3}{images/fig3/fig3.csv}

\begin{document}
\begin{filecontents*}{images/fig2/fig2.csv}
meth,colA,colB,colC,colD,colE,colF,colG,colH
,Atelectasis,Atelectasis,Effusion,Atelectasis,Atelectasis,Cardiomegaly,Atelectasis,Pneumonia
Ground Truth,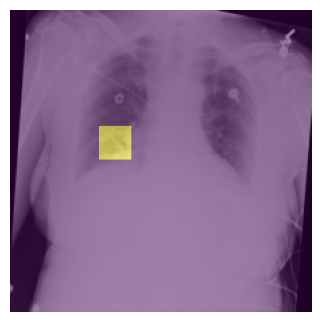,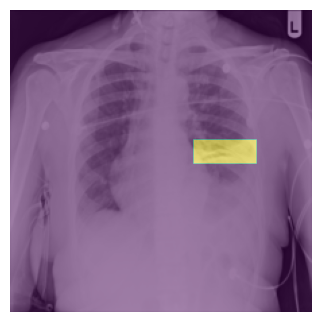,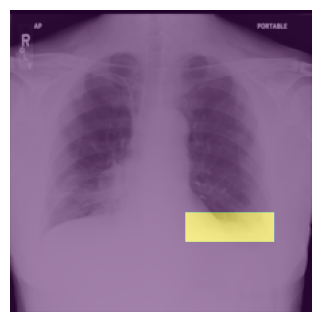,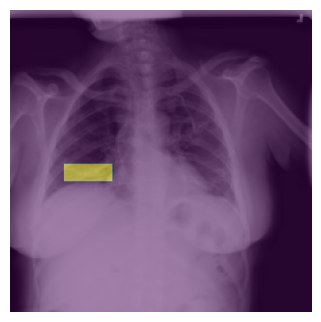,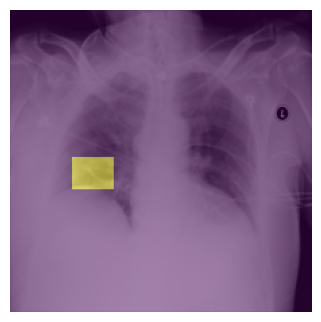,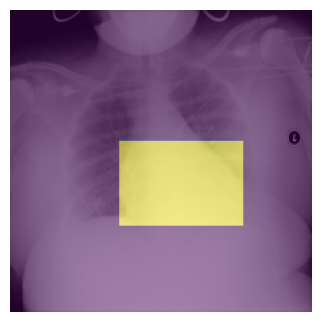,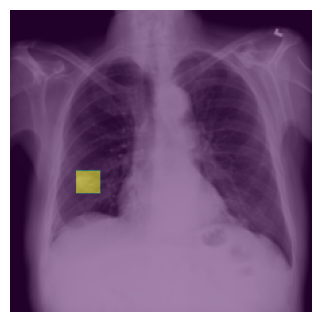,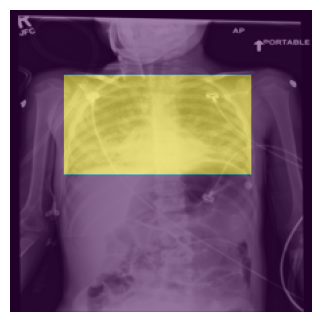
,Atelectasis,Atelectasis,Effusion,Atelectasis,Atelectasis,Cardiomegaly,Atelectasis,Pneumonia
DenseNet161 (Teacher),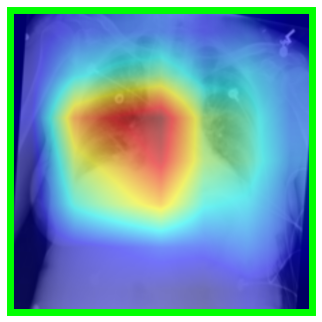,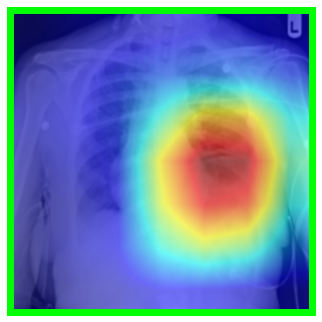,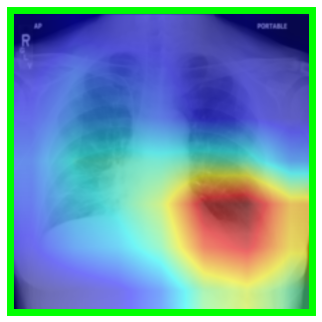,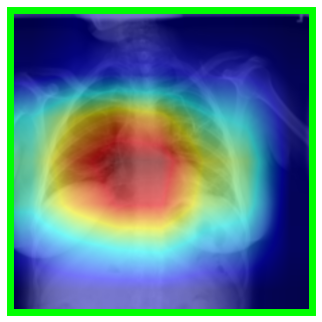,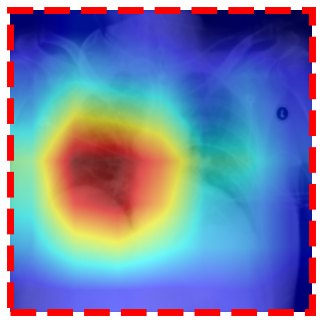,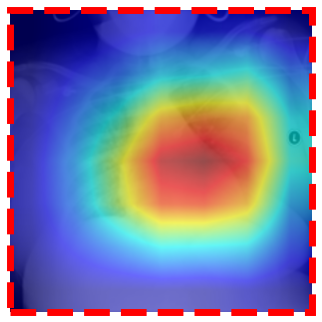,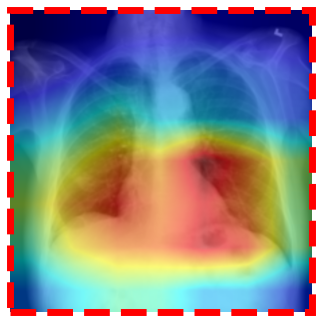,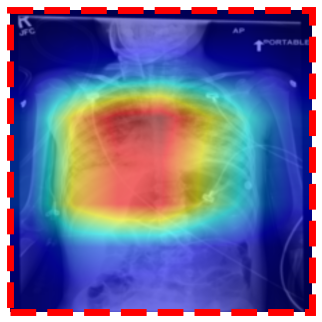
,Atelectasis,Atelectasis,Effusion,Atelectasis,Atelectasis,Cardiomegaly,Atelectasis,Pneumonia
EEEANetC2 (Student),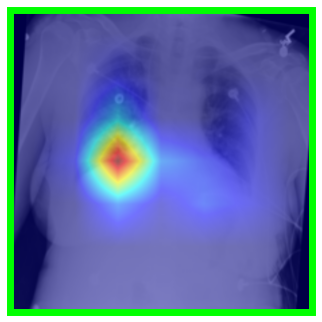,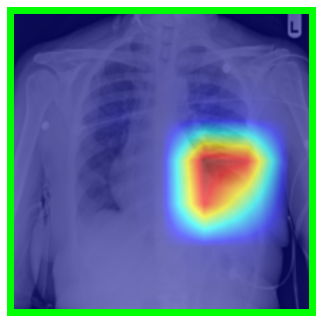,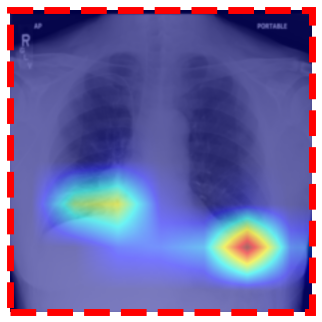,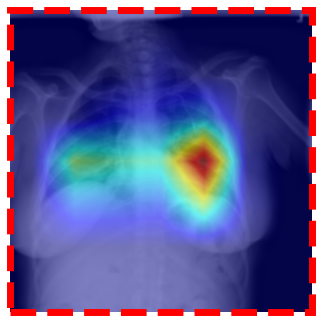,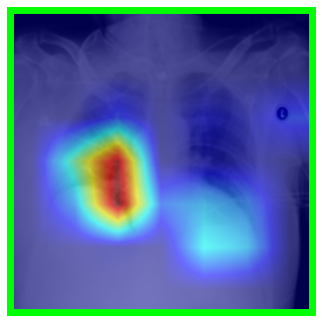,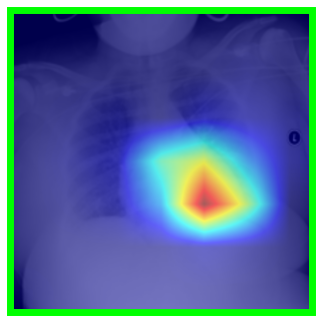,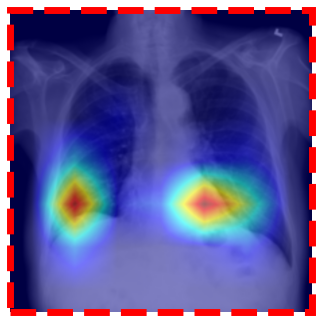,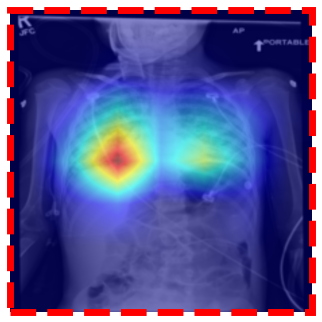
\end{filecontents*}

\begin{filecontents*}{images/fig3/fig3.csv}
meth,colA,colB,colC,colD,colE,colF
,Cardiomegaly,Infiltration,Pneumothorax,Mass,Nodule,Effusion
Ground Truth,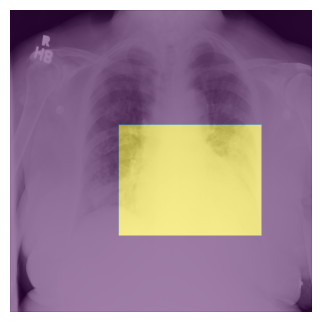,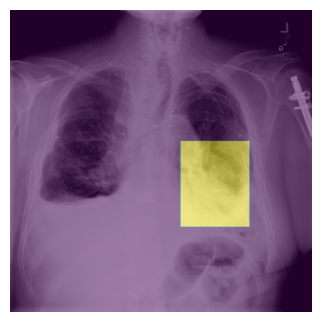,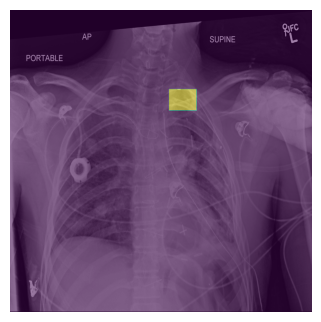,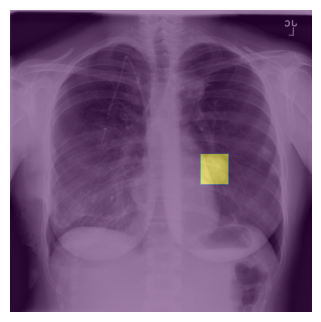,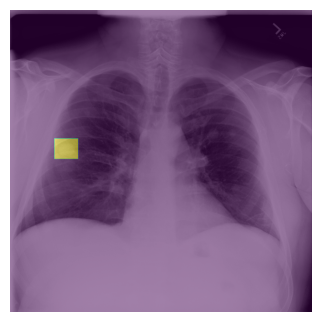,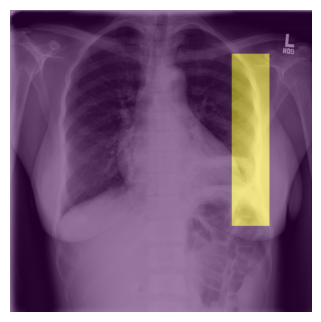
,Cardiomegaly,Infiltration,Pneumothorax,Mass,Nodule,Effusion
DenseNet161 (Teacher),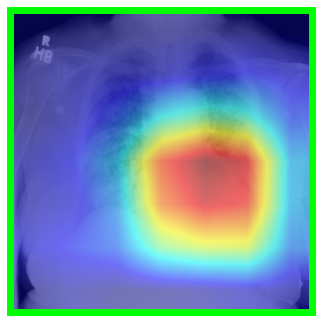,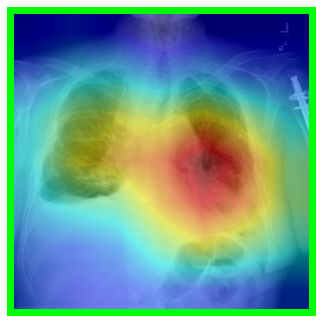,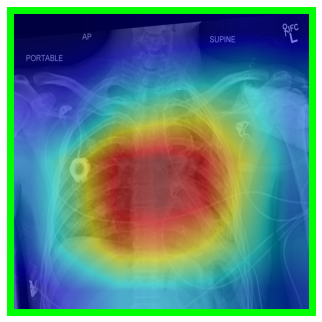,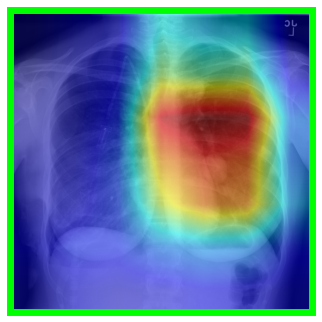,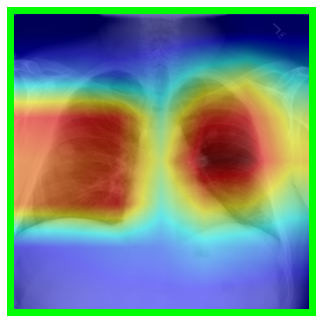,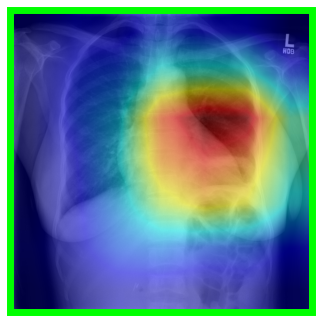
,Cardiomegaly,Infiltration,Pneumothorax,Mass,Nodule,Effusion
EEEANetC2 (Baseline),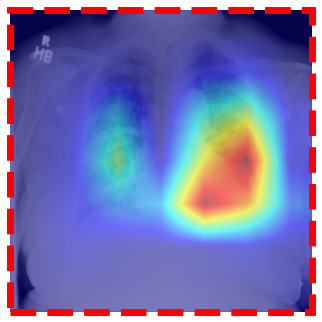,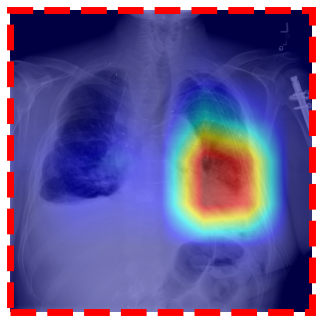,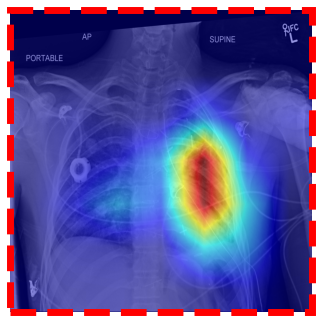,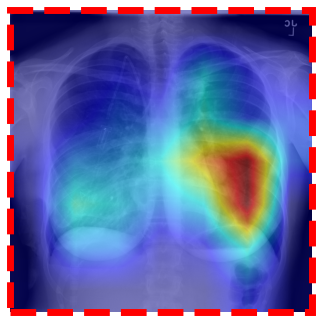,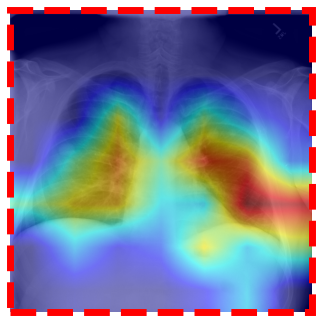,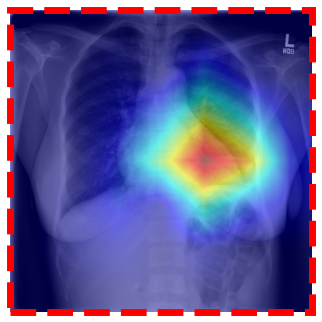
,Cardiomegaly,Infiltration,Pneumothorax,Mass,Nodule,Effusion
EEEANetC2 (Student),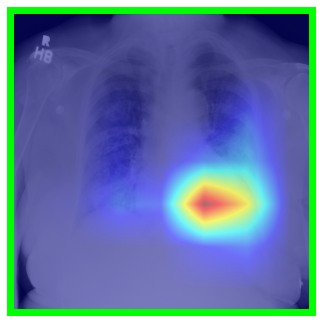,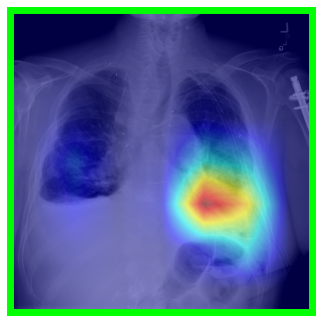,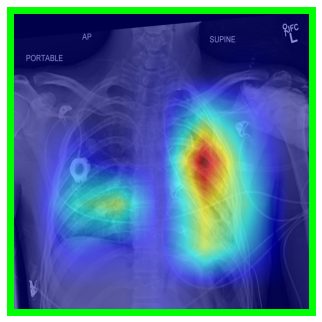,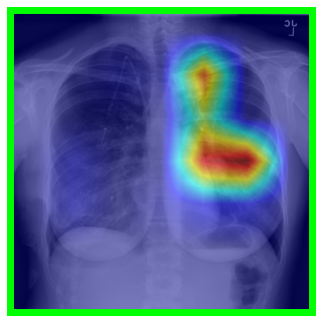,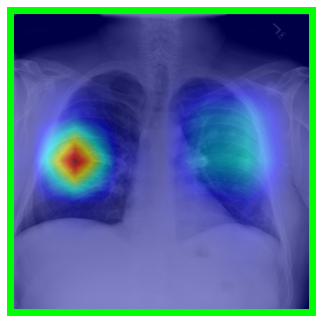,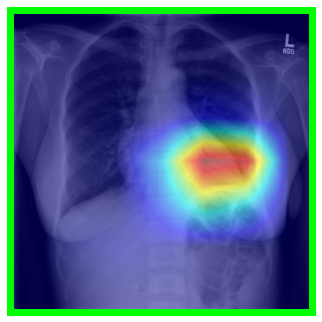
\end{filecontents*}

\title{Explainable Knowledge Distillation for On-device Chest X-Ray Classification}

\author{Chakkrit~Termritthikun,~\IEEEmembership{Member,~IEEE,}
        Ayaz~Umer,
        Suwichaya~Suwanwimolkul,~\IEEEmembership{Member,~IEEE,}
        Feng~Xia,~\IEEEmembership{Senior Member,~IEEE,} 
        and~Ivan~Lee% <-this % stops a space

% <-this % stops a space
\IEEEcompsocitemizethanks{\IEEEcompsocthanksitem C.~Termritthikun is with the School of Renewable Energy and Smart Grid Technology (SGtech), Naresuan University, Phitsanulok 65000, Thailand.\protect\\
% note need leading \protect in front of \\ to get a newline within \thanks as
% \\ is fragile and will error, could use \hfil\break instead.
E-mail: chakkritt@nu.ac.th
\IEEEcompsocthanksitem A.~Umer and I.~Lee are with the STEM, University of South Australia, Adelaide, SA 5095, Australia.\protect\\
E-mail: ayaz.umer@mymail.unisa.edu.au and ivan.lee@unisa.edu.au

\IEEEcompsocthanksitem S.~Suwanwimolkul is an independent researcher, Japan.\protect\\ E-mail: s\_suwanwimolkul@hotmail.com

\IEEEcompsocthanksitem F.~Xia is with School of Computing Technologies, RMIT University, Melbourne, VIC 3000, Australia.\protect\\ E-mail: f.xia@ieee.org

}% <-this % stops an unwanted space
% \thanks{*Corresponding author: F.~Xia}
}
% \thanks{Manuscript received April 19, 2005; revised August 26, 2015.}}

% The paper headers
\markboth{PUBLISHED AT IEEE/ACM TRANSACTIONS ON COMPUTATIONAL BIOLOGY AND BIOINFORMATICS, DOI: \url{https://doi.org/10.1109/TCBB.2023.3272333}}%
{Shell \MakeLowercase{\textit{et al.}}: Bare Demo of IEEEtran.cls for Computer Society Journals}

\IEEEtitleabstractindextext{%
\begin{abstract} Automated multi-label chest X-rays (CXR) image classification has achieved substantial progress in clinical diagnosis via utilizing sophisticated deep learning approaches. However, most deep models have high computational demands, which makes them less feasible for compact devices with low computational requirements. To overcome this problem, we propose a knowledge distillation (KD) strategy to  create the compact deep learning model for the real-time multi-label CXR image classification. We study different alternatives of CNNs and Transforms as the teacher to distill the knowledge to a  smaller student. Then, we employed explainable artificial intelligence (XAI) to provide the visual explanation for the model decision improved by the KD. Our results on three benchmark CXR datasets show that our KD strategy provides the improved performance on the compact student model, thus being the feasible choice for many limited hardware platforms. For instance, when using DenseNet161 as the teacher network, EEEA-Net-C2 achieved an AUC of 83.7\%, 87.1\%, and 88.7\% on the ChestX-ray14, CheXpert, and PadChest datasets, respectively, with fewer parameters of 4.7 million and computational cost of 0.3 billion FLOPS.

\end{abstract}

% Note that keywords are not normally used for peerreview papers.
}

% make the title area
\maketitle

% To allow for easy dual compilation without having to reenter the
% abstract/keywords data, the \IEEEtitleabstractindextext text will
% not be used in maketitle, but will appear (i.e., to be "transported")
% here as \IEEEdisplaynontitleabstractindextext when the compsoc 
% or transmag modes are not selected <OR> if conference mode is selected 
% - because all conference papers position the abstract like regular
% papers do.
\IEEEdisplaynontitleabstractindextext
% \IEEEdisplaynontitleabstractindextext has no effect when using
% compsoc or transmag under a non-conference mode.

% For peer review papers, you can put extra information on the cover
% page as needed:
% \ifCLASSOPTIONpeerreview
% \begin{center} \bfseries EDICS Category: 3-BBND \end{center}
% \fi
%
% For peerreview papers, this IEEEtran command inserts a page break and
% creates the second title. It will be ignored for other modes.
\IEEEpeerreviewmaketitle

\IEEEraisesectionheading{\section{Introduction}\label{sec:introduction}}

\IEEEPARstart{M}{edical} imaging, such as computed tomography (CT), magnetic resonance (MR), ultrasound, and X-ray, has become more important for the early identification, diagnosis, and treatment of various diseases in recent decades \cite{JACQUES202021}. The CXR is one of the most popular radiological procedures for identifying a variety of lung, heart disorder, breast cancer, and other chest-related diseases, with millions of scans conducted each year across the world \cite{england2016diagnostic,oza2022computer}. Researchers and clinicians have lately started to benefit from computer-assisted solutions due to vast variances in pathology and the potential fatigue of human experts \cite{shen2017deep,oza2022deep}.

Artificial intelligence (AI) approaches have been used in several studies on the diagnosis of chest disorders. The work can be divided into two categories, single disease classification \cite{chen2012computerized,melendez2014novel}, and multi-disease classification \cite{er2010chest,khobragade2016automatic} for CXR images. The early works analyze CXR images using traditional machine learning approaches such as support vector machines (SVMs), K-Nearest Neighbor (KNN), and Artificial Neural Networks (ANN). However, these techniques have certain limitations. Firstly, the hand-engineering features often perform well only on specific datasets. Secondly, the overfitting to certain settings due to the simplicity of model architectures. While these traditional methods are often highly efficient, the classification capability of the early machine learning is not well generalized for a wide variety of CXR images from numerous patients.

Recently, work on CXR image classification can be categorized into single \cite{wang2019enhanced} to multiple disease prediction \cite{chen2020label}. These labels represent a disorder related to the lungs, heart, or other common chest conditions. The developments in natural language processing (NLP) techniques allow radiological reports to be automatically processed to identify labels of interest for each CXR image. These enable the generation and the distribution of several CXR datasets in recent years, providing the opportunity to develop a more sophisticated model architecture, i.e., deep learning models \cite{ccalli2021deep}. The data-driven nature of deep learning benefits from large, annotated datasets, and the growing quantity of publicly available CXR imaging datasets enables the building of highly accurate models \cite{rehman2021survey}. Due to the superior performance of deep learning, it has become highly attractive for medical image analysis \cite{Zhou2021SODA,wang2019enhanced,Song2021DeepLearning,Pathak2021DeepBidirectional,zhang2021dynamic}. Deep neural networks (DNNs) such as DenseNet ~\cite{huang2017densely} provide state-of-the-art accuracy in medical image applications.

Nevertheless, they also present challenges for deployment on the smaller devices due to their computing cost for the dense features. Vision Transformer \cite{dosovitskiy2021an} is another state-of-the-art, whose accuracy is gained by their self-attention mechanism. Due to its high time and memory costs, self-attention is often avoided for many on-device and real-time applications. In the case of real-time applications, there is a need for models that can perform quick inference. Consequently, smaller, and more compact DNNs with fewer parameters and lower computational complexity are preferred. These models require minimal processing complexity and can provide faster inference times in real-time applications. Using smaller DNNs such as \cite{howard2019searching,termritthikun2021eeea} can reduce such costs, yet their accuracy is much lower compared to the large DNNs. Thus, there is a demand for a compact DNN, which is computationally efficient and able to keep the same high level of performance for on-device deployment and real-time CXR image analysis.

To address this problem, we explored a teacher-student technique known as KD \cite{umer2022ondevice} to create a compact student model with less computational complexity. Knowledge of a teacher model is utilized to boost the performance of the student model in the teacher-student framework. KD has recently been used in several CXR image classification works. For instance, numerous works \cite{ho2020untilizing,sonsbeek2021variational,Chen2021multilabelchest,li2020covid} can be seen in CXR image classification domain. Furthermore, XAI is a new machine learning research topic that aims to unbox how AI systems make black-box decisions. XAI is a term that refers to AI tools that can explain the deep learning operation to be comprehensible to end-users \cite{arrieta2020explainable}. Recently, XAI has gained popularity and attracted significant interest from the medical image analysis research community, as it tackles two essential issues: transparency and accountability \cite{Tjoa2021asurvey}. CXR image classification \cite{ho2020untilizing,sonsbeek2021variational,Chen2021multilabelchest,wang2019enhanced} utilize XAI for explaining their models to boost the confidence. The preceding research articles use visualization-based XAI techniques to give visual explanations for their models.
 
We propose a KD strategy where the teacher and student are of different types of architecture. In addition, we provide an in-depth study on the interpretability in AI to explain the decision resulting from the complex processing inside deep learning with our proposed KD strategy. Our contributions are:

\begin{itemize} 
\setlength\itemsep{0.5em}
   
\item We proposed to transfer the knowledge from architectures where the student and teachers are of different types, i.e., from Transformer/CNN-based teacher to a CNN-based student model. Our study considers various SOTA image classification models based on CNN (DenseNet~\cite{huang2017densely}, Once-For-All (OFA)~\cite{cai2020once}, and Transformer (AutoFormer~\cite{chen2021autoformerv2}, Visformer\cite{chen2021visformer}) to build a lightweight multi-label CXR image classification model.
  \item We incorporated the XAI strategy to establish the trust and transparency in the proposed model, where the end-user, e.g., a radiologist, can receive the visual explanations in the form of attention heat maps. The attention heat map provides the explanation for the disease prediction patterns that student learns from the teachers.          
  \item Our student model provides higher AUC compared to the baseline, additionally our learning also enables the accurate knowledge transfer from the teacher to student network. Furthermore, XAI visualizations in the form of heatmaps show a strong correlation between the annotation provided by the radiologist and student model. 
 \end{itemize}

\section{Related work}
In this section, we provide a comprehensive review of the related work on CXR image classification and utilization of XAI in this domain. Furthermore, we also provided a thorough analysis of KD approaches and On-device methods utilized for CXR images.

\subsection{Deep learning for CXR Images}
Machine learning techniques have been utilized in exploring automatic diagnosis for CXR images to ease the heavy workload of radiologists. These techniques enable the development of systems capable of predicting more than one disease by utilizing a multi-label classification strategy for CXR images. The work \cite{er2010chest} utilized an artificial neural network to diagnose five important chest diseases using CXR images. They conducted a comparative study using multilayer, probabilistic learning vector quantization, and regression neural networks and reported the average classification of each disease for a single dataset. Another work \cite{khobragade2016automatic} proposes a system for CXR images that diagnoses three important chest-related diseases by combining the segmentation and classification tasks. They improved the performance by using histogram equalization and pattern recognition techniques, with the limitation that the size and position of the CXR image be fixed. The work \cite{ahmad2014effectsof} focus on detecting infectious region in CXR images by utilizing various classifiers, including Bayesian, k-nearest neighbor's, and Rule-based. They have evaluated their technique using various assessment criteria, including classification accuracy, speed, the Kappa statistic, misclassification (error), the receiver operating characteristic (ROC) curve, precision, and recall.

In recent years there has been an increase in techniques exploring automated diagnosis for radiological images. Researchers are increasingly using data-driven approaches, such as deep learning, to analyze chest CXR images to enhance efficiency and ease the burden on radiologists. This progressively increasing trend is due to the availability of large publicly available datasets, including ChestXray14 \cite{Wang2017ChestXRay8}, CheXpert \cite{Irvin2019CheXpert}, and PadChest \cite{Aurelia2020PadChest}. Recently some interesting work for CXR images analysis has been published \cite{Zhou2021SODA,Song2021DeepLearning,chen2020label,Chen2021multilabelchest}, which shows the effectiveness of the DenseNet \cite{huang2017densely} for CXR image analysis. DenseNet is based on the neural network design used for visual recognition. In DenseNet architecture, each layer is connected to every other layer in a feed-forward fashion. The feature maps in the current layer are concatenated with those from all the preceding layers, thus providing rich feature representation. Furthermore, the popularity of DenseNet may be attributed to a variety of convincing factors, including enhanced feature propagation, the ability to handle the vanishing-gradient problem, and fewer parameters.

Recently, Vision Transformer (ViT) \cite{dosovitskiy2021an} has received popularity due to its superior capability to capture long-range dependencies. The attention-based architectures outperform CNNs by capturing global information as the small kernel size of CNNs can limit their receptive field. On the other hand, Transformers can capture long-range dependencies due to the self-attention mechanism integrated into their architecture. Vision Transformer-based architectures are getting more attention from researchers for CXR image analysis. For instance, xViTCOS \cite{mondal2021xvitcos} is a Vision Transformer-based architecture with a multi-stage transfer learning method to address the data scarcity problem. The evaluation results show that it outperforms state-of-the-art models over various evaluation metrics for COVID-19 screening. Another work \cite{park2022multi} proposed an algorithm based on a Vision Transformer for COVID-19 diagnosis to quantify the severity of the virus disease. They utilize a multi-task approach that leverages low-level CXR features corpus called embedding features obtained from a backbone network. These embedding features are used as corpora for Transformers.

\subsection{Knowledge Distillation for CXR Images}
KD has recently been used in several CXR image classification works. For instance, a competitive study of KD is presented \cite{ho2020untilizing} for classifying abnormalities in CXR images, where they evaluated their method on a single dataset by considering CNN-based architectures. Another work \cite{sonsbeek2021variational} proposed a variational knowledge distillation (VKD) which is a probabilistic inference framework for the disease classification on CXR images, and the KD uses the knowledge from Electronic Health Records (EHR) in learning. Experienced radiologists can establish clinical diagnoses based on correlations between various CXR images. Inspired by this, Semantic Similarity Graph Embedding (SSGE) \cite{Chen2021multilabelchest} built on the "Teacher-Student" (semantic-visual) learning technique, explores the semantic similarities among images for improving the performance of multi-label CXR image classification. COVID-MobileXpert \cite{li2020covid} is proposed as a lightweight deep neural network for COVID-19 screening and radiological trajectory prediction. The proposed method creates a hardware-friendly student model that targets COVID-19 screening by utilizing Knowledge transfer and distillation (KTD).

\subsection{Explainable AI for CXR Images}
Recent studies on XAI aim to make AI models more interpretable to increase the trust in intelligent systems and bring them closer to meaningful integration into ordinary life \cite{arrieta2020explainable}. Despite its great achievements in the medical domain, AI-based technologies have yet to attain widespread adoption in clinics due to the inherent black-box nature of deep learning. The medical diagnostic system must be transparent, understandable, and explainable to earn the confidence of physicians, regulators, and patients \cite{Tjoa2021asurvey}. Recently, XAI has gained popularity in the research community for CXR image analysis. Heatmap explanation is utilized, including Class Activation Map (CAM) \cite{zhou2016learning} and SmoothGrad integrated Gradients to provide the visual explanation of a proposed method \cite{ho2020untilizing}. In this work, Smoothgrad aims to reduce noise and visual diffusion, while CAM is used to extract weight activations from convolutional layers. A work proposed \cite{sonsbeek2021variational} KD to acquire knowledge from EHRs to improve their model performance. They integrate XAI to provide a visual explanation of their method by utilizing Gradient-weighted class activation mapping (Grad-CAM) \cite{selvaraju2017gradcam} by comparing the importance of words from EHR and CXR images. Another work \cite{Chen2021multilabelchest} proposed a novel Semantic Similarity Graph Embedding (SSGE) framework, exploring the semantic similarities between CXR images for multi-label classification. They provide visual explanations by utilizing Grad-CAM to localize the lesion area by comparing the manual lesion regions with the Grad-CAM heatmaps corresponding regions. Another study \cite{wang2019enhanced} used deep convolutional neural networks (DCNN) to identify a common lung condition called pneumothorax. It used Grad-CAM to construct thermodynamic diagrams of CXR images to display the lesion location properly.

\subsection{On-device methods for CXR Images}
Creating an accurate and computationally efficient model design is a critical yet difficult aspect of developing any high-performance machine learning system. Conventional approaches either manually design or use neural architecture search (NAS) to find a specialized neural network and train it from scratch for each case, which is computationally prohibitive because it requires high computation \cite{liu2021survey}. However, a recent approach has emerged in a supernet form that assembles all candidate architectures into weight sharing network with each architecture corresponding to one subnet. Various architectures can directly inherit the weights from the supernet for evaluation and deployment, which eliminates the huge cost of training. For instance, OFA \cite{cai2020once} is a new algorithm that supports a wide range of architectural settings and creates many subnetworks adaptive to a wide range of hardware platforms and latency constraints while maintaining the same degree of accuracy as training individually.

Designing a Transformer to realize high efficiency is a difficult task. The network depth, the embedding size, and the number of heads can significantly impact Vision Transformer performance. These dimensions had previously been required to be explicitly set manually, which was a tedious task. An AutoFormer \cite{chen2021autoformerv2} can be utilized to automate the Transformer design process. In addition, a novel design called Visformer \cite{chen2021visformer} is presented, which is a vision-friendly Transformer. Visformer outperforms both Transformer- and convolution-based models in ImageNet classification with the same computational complexity.

On-device CXR image analysis has attracted much interest, but more research is needed to get the most out of low-computation models for CXR image analysis. For instance, a work \cite{li2020covid} addressed COVID-19 screening by proposing a lightweight deep neural network specifically designed for mobile phone systems. Even though their model is small, it is platform-specific and only considers one disease, i.e., COVID-19. Another research \cite{masud2022light} utilized CXR images to create a compact, lightweight CNN model for COVID-19 disease classification, yet their model is designed for binary classification.

%  Fig. 1
\begin{figure*}[t!]
  \begin{center}
    \includegraphics[width=0.85\linewidth]{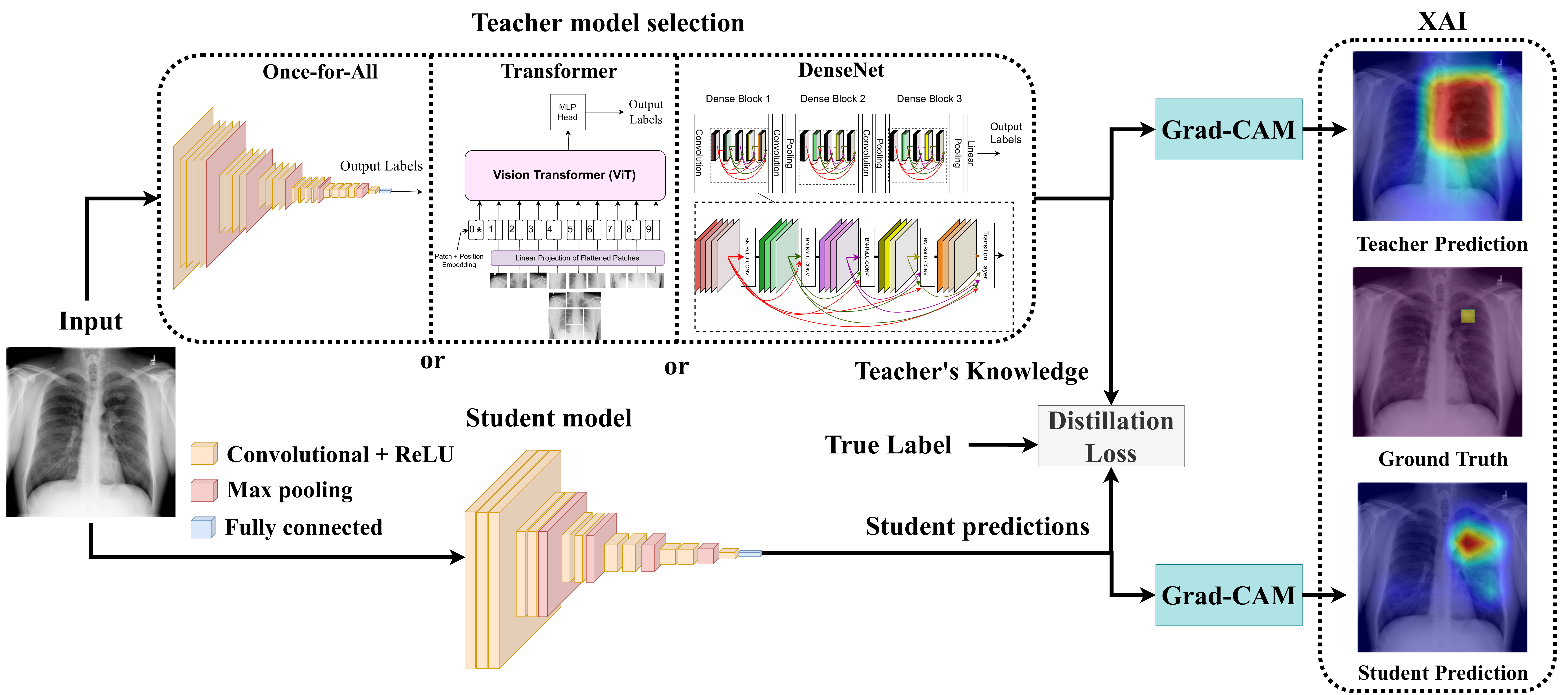}
  \end{center}
  \caption{The overall architecture of the proposed on-device CXR prediction model. Proposed method transfer knowledge from teacher to student model and utilize Grad-CAM to provide visual explanations for model decision. For training, we choose one teacher at a time from a pool of five teachers, namely OFA-595, DenseNet161, EEEA-Net-C2, Visformer-S, and AutoFormerV2-T.}
  \label{fig:1}
\end{figure*}

\section{Method}
The process of KD permits the transfer of information from a larger model to a smaller one. However, there are significant challenges in optimizing models for on-device CXR image classification, for instance, constraints in building smaller models for devices with limited resources, medium size dataset, type of images are grayscale, and class imbalance problem. Therefore, we proceed with the following steps to address the problems. 

\subsection{Framework}

The explainable KD method for creating on-device CXR classification is shown in Fig.~\ref{fig:1}. It consists of three major components: a large network (teacher) and a small network (student), distillation loss, and explainable model behavior using visualizations. To improve performance, the on-device CXR prediction model employs KD to teach a small model to emulate a larger pre-trained model.

The total loss of the student model is the combination of its hard and soft losses. The proposed on-device CXR prediction method is based on multi-label classification, where it takes input as a grayscale CXR image of size  $W \times H$  and generates a predicted label. The teacher network is trained initially using the CXR image dataset, and then the student network is trained by guiding the training of a student model using predictions from a teacher network.

\subsection{On-device CXR Classification Architectures}

The proposed method, on-device CXR image classification, aims to minimize computational complexity in a multi-label classification problem by distilling knowledge from a teacher network to a student network. Multiple cutting-edge teacher networks are selected and then utilized to train a student model. OFA-595, DenseNet161, EEEA-Net-C2, Visformer-S, and AutoFormerV2-T models are used as teacher networks, while the EEEA-Net-C2 model is used as a student network. Furthermore, based on the architectural design, we divided teacher networks into two categories: CNN-based teachers and Transformer-based teachers.

%  Algo. 1
\begin{algorithm}[!t]
    \caption{Knowledge Distillation (KD)}
    \label{alg:1}
    \begin{algorithmic}[1]
        \STATE {\bfseries Input:} Teacher $T$ (trained with chest x-ray images), Initialized Student $S$.
        \STATE {\bfseries Output:} Trained Student $S$
        \STATE $p_t$ is the prediction of teacher, $p_s$ is the prediction of student, $y$ is the ground truth label, $\alpha$ is the weight of hard loss and soft loss, $\mathcal{L}_{(\cdot)}$ is the associated loss function, $D$ is the training data.

    \FOR {each batch in $D$}
       \STATE $p_t$ $\gets$ Forward $T$ given input images;
       \STATE $p_s$ $\gets$ Forward $S$ given input images; 
       %\STATE $FL \gets L_{T}, MSE \gets L_{S} $
      \STATE Calculate $\mathcal{L}_{FBCE}$ with $p_s, y$; (see Eq.~\ref{eq:4})
      \STATE Calculate $\mathcal{L}_{MSE}$ with $p_s, p_t$; (see Eq.~\ref{eq:5})
      \STATE $\mathcal{L}_{KD}$  $\gets$ $\alpha \cdot \mathcal{L}_{FBCE} + (1-\alpha) \cdot \mathcal{L}_{MSE}$;
      \STATE $S$ $\gets$ Backward  by $\mathcal{L}_{KD}$ 
      \ENDFOR
      \STATE Return $S$
    \end{algorithmic}
\end{algorithm}

\subsubsection{CNN-based Teachers}

The OFA supernet model \cite{cai2020once} is trained once and then specialized for efficient deployment by allowing it to be deployed in a variety of architectural configurations. It uses a novel progressive shrinking algorithm to shrink the model across multiple dimensions (depth, width, kernel size, and resolution). Furthermore, it can generate numerous subnetworks that are compatible with a wide range of hardware platforms and latency constraints while maintaining high accuracy. We chose OFA-595 as a subnet from the OFA supernet due to its superior accuracy.

According to the author of EEEA-Net-C2 \cite{termritthikun2021eeea}, it is a subnet of the OFA supernet. To increase performance, we also utilized the approach of using the same teacher and student model for KD. We chose EEEA-Net-C2 as the teacher and student model, and by utilizing the same student and teacher network, we were able to increase the student model's performance.

DenseNet161 was chosen as the teacher model due to the widespread use of DenseNet in CXR image processing, making it a viable candidate for selection. DAM \cite{yuan2021large}, which is based on DenseNet-121 architecture, claims that the reported results achieved state-of-the-art of the CheXpert competition in 2020.

\subsubsection{Transformer-based Teachers}

ViT is based on a self-attention mechanism, while CNN works on convolutions both have different architectures and working principles, thus CNN-based model will acquire different knowledge compared to Vision Transformer-based architecture on the same dataset.

Several works in the literature are derived from ViT to improve performance for a variety of applications. The proposed work considered recent ViT-based architectures by selecting the lightweight version of NAS-based transformer (Visformer-S\cite{chen2021visformer}, and AutoFormer-V2-T\cite{chen2021autoformerv2}), which are both efficient and compact models when compared to the original Vision Transformer-based model. These architectures are designed specifically for natural image classification, the proposed method adapted these architectures for CXR multi-label image classification.

\subsection{CXR Distillation Loss}
The conventional approach of employing KD to train multi-label classification networks is proposed, in which predictions from a teacher network guide the training of a student model. The objective is to learn from a teacher model using a student model while maintaining a reasonable degree of accuracy.

A multi-label classification problem generally gives an output layer with sigmoid activation between 0 and 1. The probability scores and true label are fed into the Binary Cross-Entropy (BCE) loss function to calculate the loss value, which can be computed by using the following equation:

% Eq.  1
\begin{equation}
  \label{eq:1}
   \mathcal{L}_{BCE}(p,y) = \left\{ \begin{array}{ll} -\log(p), & \mathrm{if} \ y=1 \\ -\log(1-p), & \mathrm{otherwise} \end{array} \right.
\end{equation}
\noindent where $p$ is the probability of prediction score, and $y$ is the ground truth.
 Let $p_s$ denote the probability of student prediction score. \eqref{eq:1} can be rewritten as: 
\begin{equation}
  \label{eq:2}
  p_s = \left\{ \begin{array}{ll} p, & \mathrm{if} \ y=1 \\ 1-p, & \mathrm{otherwise} \end{array} \right.
\end{equation}
\noindent  Thus, the $\mathcal{L}_{BCE}$ becomes $\mathcal{L}_{BCE}(p,y)=\mathcal{L}_{BCE}(p_{s})=-\log(p_{s})$.
% % Eq.  3
% \begin{equation}
%   \label{eq:3}
%   \mathrm{BCE}(p,y)=\mathrm{BCE}(p_{s})=-\log(p_{s})
% \end{equation}
 
The class imbalance problem in multi-label CXR classification is a critical task since it directly affects the model's efficiency. To overcome this problem, we introduce Focal Binary-Cross Entropy loss ($FBCE$), which adopts Focal loss ($FL$) \cite{lin2017focal} for BCE loss with multi-label CXR problem. FL focuses on the hard examples (samples whose predicted probability is much lower than 1) by expanding the criteria of well-classified examples by increasing value of $\gamma$, to improve the predictions on hard examples over the time rather than becoming overconfident with easy examples.

% Eq.  4
\begin{equation}
  \label{eq:4}
  \mathcal{L}_{FBCE}(p_{s},y)=-(1-p_{s})^{\gamma}\log(p_{s})
\end{equation}

\noindent $(1-p_{s})$  is the modulating factor in Eq.~\ref{eq:4}. When a model predicts a higher confidence score, the modulating factor will tend to become closer to 0; thus, down-weighting the loss value for well-classified examples. $\gamma$ is the tunable hyperparameter, which adjusts how much easy examples need to be down-weighted, its value ranging from 1-5\cite{lin2017focal}. If $\gamma$ is 0, the FBCE loss is equal to the BCE loss.

Conversely, $MSE$ loss employs soft loss through teacher prediction and student prediction scores, for the multi-label problem, we applied the sigmoid activation function to compute both prediction scores into a value between 0 and 1, which can be computed by using the following equation:

% Eq.  5
\begin{equation}
  \label{eq:5}
\mathcal{L}_{MSE}(p, q)=T^{2}\cdot  \frac{1}{n}\sum_{i=1}^{n} \frac{\left ( p_{i} - q_{i} \right )^{2}}{T} 
\end{equation}
 
\noindent where $T$ represents temperature factor control soft targets, $p$ represents the student model prediction score computed by $\mathrm{Sigmoid}(p_i/T)$, and $q$ represents the teacher model prediction score computed by $\mathrm{Sigmoid}(q_i/T)$.

The proposed on-device CXR prediction method utilizes two losses, including $FBCE$ and $MSE$ losses. Total loss is the weighted sum of the $FBCE$ hard loss and $MSE$ soft loss. Finally, the student network is trained to optimize the total loss given by using the following equation: 
% Eq.  6
\begin{equation}
  \label{eq:6}
  \mathcal{L}_{KD}=\alpha \cdot \mathcal{L}_{FBCE}(p_s,y)+\left ( 1-\alpha \right ) \cdot \mathcal{L}_{MSE}(p_{s},p_{t})
\end{equation}

\noindent where $p_s$ and $p_t$ indicate the student and teacher prediction score, respectively. 
$y$ represents ground truth, and $\alpha$ represents the weights for controlling hard and soft losses. In this paper, we observed that weights value $\alpha=0.5$ same as traditional KD configuration to balance out hard and soft losses. For more detail, see Algorithm~\ref{alg:1}. 

\subsection{Explainable AI for Knowledge Distillation}
In this section, we chose the Grad-CAM method \cite{selvaraju2017gradcam} as the visual representation method. It is a class-discriminative localization approach that generates heatmap-based visual explanations for any CNN-based network without needing architectural modifications or re-training. Grad-CAM can be used to acquire a decent comprehension and understanding of the model prediction when transferring teacher knowledge to the student model.

We utilized Grad-CAM method for explaining our proposed model's decision by using visual explanations, mathematically, the class-discriminative localization map is represented by $ L^c_{Grad-CAM}\epsilon\emph {R}^{m\times n}$ of width $m$ and height $n$ for any class $c$. The first step is to compute the gradient of the score for class $c$, which is represented by the gradient with respect to activations of the feature map $A^k$ of a convolutional layer represented by $\frac{\partial y^c}{\partial A^k}$. Finally, the neuron significance weights are calculated by computing the global average pooling of the gradients flowing back across the width and height dimensions($i$ and $j$). 

% $ \alpha^{c}_{k} $

% Eq.  7
\begin{equation}
  \label{eq:7}
\alpha^{c}_{k}=\frac{1}{Z}\sum_i \sum_j \frac{\partial y^c}{\partial A^k_{i_j}}
\end{equation}

\noindent The weights $\alpha^{c}_{k}$ represent a partial linearization of deep network downstream from $A$, and captures the importance of feature map $k$ for a target class $c$. Finally, a weighted combination of forward activation maps followed by $ReLU$ is represented by 

% Eq.  8
\begin{equation}
  \label{eq:8}
L^c_{Grad-CAM}=ReLU(\sum_k\alpha^c_k A^k)
\end{equation}

\begin{table*}[!t]
\centering
\caption{Baseline performance, number of parameters and computational complexity of EEEA-Net-C2, OFA-595, DenseNet161, Visformer-S, and  AutoFormerV2-T over three datasets including ChestX-ray14, CheXpert and PadChest.}
\label{tab:1}
\begin{tabular}{l|ccc|ccc|ccc|cc}
\hline 
\textbf{Datasets} & \multicolumn{3}{c|}{\textbf{ChestX-ray14}~\cite{Wang2017ChestXRay8}} & \multicolumn{3}{c|}{\textbf{CheXpert}~\cite{Irvin2019CheXpert}} & \multicolumn{3}{c|}{\textbf{PadChest}~\cite{Aurelia2020PadChest}} & \textbf{} & \textbf{} \\ \hline
\textbf{Model} & \textbf{\begin{tabular}[c]{@{}c@{}}AUC\\ (\%)\end{tabular}} & \textbf{\begin{tabular}[c]{@{}c@{}}Accuracy\\ (\%)\end{tabular}} & \textbf{\begin{tabular}[c]{@{}c@{}}F1\\ (\%)\end{tabular}} & \textbf{\begin{tabular}[c]{@{}c@{}}AUC\\ (\%)\end{tabular}} & \textbf{\begin{tabular}[c]{@{}c@{}}Accuracy\\ (\%)\end{tabular}} & \textbf{\begin{tabular}[c]{@{}c@{}}F1\\ (\%)\end{tabular}} & \textbf{\begin{tabular}[c]{@{}c@{}}AUC\\ (\%)\end{tabular}} & \textbf{\begin{tabular}[c]{@{}c@{}}Accuracy\\ (\%)\end{tabular}} & \textbf{\begin{tabular}[c]{@{}c@{}}F1\\ (\%)\end{tabular}} & \textbf{\begin{tabular}[c]{@{}c@{}}Params\\ (M)\end{tabular}} & \textbf{\begin{tabular}[c]{@{}c@{}}FLOPS\\ (G)\end{tabular}} \\ \hline
EEEA-Net-C2 \cite{termritthikun2021eeea} & 81.0 & 93.8 & 93.2 & 85.8 & 85.8 & 83.7 & 85.8 & 96.8 & 96.3 & \textbf{4.7} & \textbf{0.3} \\
OFA-595 \cite{cai2020once} & 82.6 & \textbf{94.1} & \textbf{93.4} & 86.7 & \textbf{86.5} & \textbf{84.5} & 88.0 & 96.7 & 96.4 & 7.6 & 0.5 \\
DenseNet161 \cite{huang2017densely} & \textbf{83.8} & 93.9 & \textbf{93.4} & 87.0 & 86.2 & 84.2 & 88.9 & \textbf{96.9} & \textbf{96.6} & 26.5 & 7.8 \\
Visformer-S \cite{chen2021visformer} & 82.9 & 93.9 & \textbf{93.4} & \textbf{87.4} & 86.2 & \textbf{84.5} & \textbf{89.1} & 96.8 & 96.5 & 39.5 & 4.8 \\
AutoFormerV2-T \cite{chen2021autoformerv2} & 82.3 & \textbf{94.1} & \textbf{93.4} & 86.7 & 86.0 & 83.9 & 87.4 & 96.8 & 96.4 & 27.6 & 4.4 \\ \hline
\end{tabular}
\end{table*}

\begin{table*}[t]
\centering 
\caption{Performance of the proposed strategy (Knowledge Distillation), where different teachers are used for training the student EEEA-Net-C2. DenseNet161 outperformed other teacher models over various evaluation metrics (AUC, Accuracy, F1) across three datasets including ChestX-ray14, CheXpert and PadChest.} 
\label{tab:2}
\begin{tabular}{lccccccccccc}
\hline
\multicolumn{3}{l}{\textbf{Datasets}} & \multicolumn{3}{c}{\textbf{ChestX-ray14}~\cite{Wang2017ChestXRay8}} & \multicolumn{3}{c}{\textbf{CheXpert}~\cite{Irvin2019CheXpert}} & \multicolumn{3}{c}{\textbf{PadChest}~\cite{Aurelia2020PadChest}} \\ \hline
\textbf{Method} & \textbf{Teacher} & \multicolumn{1}{c|}{\textbf{Student}} & \textbf{\begin{tabular}[c]{@{}c@{}}AUC\\ (\%)\end{tabular}} & \textbf{\begin{tabular}[c]{@{}c@{}}Accuracy\\ (\%)\end{tabular}} & \multicolumn{1}{c|}{\textbf{\begin{tabular}[c]{@{}c@{}}F1\\ (\%)\end{tabular}}} & \textbf{\begin{tabular}[c]{@{}c@{}}AUC\\ (\%)\end{tabular}} & \textbf{\begin{tabular}[c]{@{}c@{}}Accuracy\\ (\%)\end{tabular}} & \multicolumn{1}{c|}{\textbf{\begin{tabular}[c]{@{}c@{}}F1\\ (\%)\end{tabular}}} & \textbf{\begin{tabular}[c]{@{}c@{}}AUC\\ (\%)\end{tabular}} & \textbf{\begin{tabular}[c]{@{}c@{}}Accuracy\\ (\%)\end{tabular}} & \textbf{\begin{tabular}[c]{@{}c@{}}F1\\ (\%)\end{tabular}} \\ \hline
Baseline & - & \multicolumn{1}{c|}{EEEA-Net-C2~\cite{termritthikun2021eeea}} & 81.0 & 93.8 & \multicolumn{1}{c|}{93.2} & 85.8 & 85.8 & \multicolumn{1}{c|}{83.7} & 85.8 & 96.8 & 96.3 \\ \hline
KD & EEEA-Net-C2~\cite{termritthikun2021eeea} & \multicolumn{1}{c|}{EEEA-Net-C2~\cite{termritthikun2021eeea}} & 82.5 & \textbf{94.3} & \multicolumn{1}{c|}{93.5} & 87.0 & 86.2 & \multicolumn{1}{c|}{84.4} & 87.0 & 96.9 & 96.4 \\
KD & OFA-595~\cite{cai2020once} & \multicolumn{1}{c|}{EEEA-Net-C2~\cite{termritthikun2021eeea}} & 83.4 & 94.2 & \multicolumn{1}{c|}{93.5} & 87.3 & 86.7 & \multicolumn{1}{c|}{84.7} & 88.4 & 96.7 & 96.5 \\
KD & DenseNet161~\cite{huang2017densely} & \multicolumn{1}{c|}{EEEA-Net-C2~\cite{termritthikun2021eeea}} & \textbf{83.7} & 93.9 & \multicolumn{1}{c|}{93.4} & 87.1 & \textbf{86.8} & \multicolumn{1}{c|}{\textbf{84.9}} & 88.7 & \textbf{97.0} & \textbf{96.6} \\
KD & Visformer-S~\cite{chen2021visformer}  & \multicolumn{1}{c|}{EEEA-Net-C2~\cite{termritthikun2021eeea}} & 83.1 & 94.1 & \multicolumn{1}{c|}{93.5} & \textbf{87.6} & 86.4 & \multicolumn{1}{c|}{84.5} & \textbf{88.8} & \textbf{97.0} & \textbf{96.6} \\
KD & AutoFormerV2-T~\cite{chen2021autoformerv2} & \multicolumn{1}{c|}{EEEA-Net-C2~\cite{termritthikun2021eeea}} & 83.3 & \textbf{94.3} & \multicolumn{1}{c|}{\textbf{93.6}} & 87.2 & 86.4 & \multicolumn{1}{c|}{84.5} & 87.9 & \textbf{97.0} & 96.5 \\ \hline
\end{tabular}
\end{table*}

\noindent where $k$ is the index of the activation map, and $c$ is the class of interest. Finally, we multiply each activation map $A^k$ by its importance score $\alpha^{c}_{k}$ and pass it through the $ReLU$ function to only consider the pixels that contribute a positive influence on the score of the class of interest.

\section{Experimental Results}

\subsection{Datasets}

We employed three public datasets, \ie,  ChestX-ray14~\cite{Wang2017ChestXRay8}, CheXpert~\cite{Irvin2019CheXpert}, and PadChest~\cite{Aurelia2020PadChest}, to provide the performance evaluation and the analysis.

\smallskip 
\noindent $\bullet$ ChestX-ray14 contains 112120 frontals CXR images from 30805 individuals with the text-mined fourteen common disease labels, mined from the text radiological reports via NLP techniques. The images are labeled with 14 diseases, where each image can
have multi-labels.

\smallskip 
\noindent $\bullet$ CheXpert contains 224316 CXR images from 65240 Stanford Hospital inpatients and outpatients between October 2002 and July 2017. The participants are primarily adults whose CXR images are automatically labeled according to the radiology reports using a rule-based labeler, identifying the presence, absence, or uncertainty of 12 abnormalities, no-findings, and existence of support devices.

\smallskip
\noindent $\bullet$ PadChest contains 158626 CXR images from 67000 patients, interpreted and reported by radiologists at Hospital San Juan Hospital from 2009 to 2017. The CXR images are labeled with 19 differential diagnoses. Twenty-seven percent of them were manually annotated by trained physicians, while the rest were labeled with a supervised method based on a recurrent neural network~\cite{Aurelia2020PadChest}.

\subsection{Evaluation Metrics}
Following~\cite{chen2020label,Chen2021multilabelchest}, we utilize the area under the ROC curve (AUC) to evaluate the disease prediction performance in multi-label classification. Additionally, accuracy and F-score metrics are also used for comprehensive evaluation.

\begin{table*}[t]
\centering
\caption{Label-wise (14 labels) baseline performance of teachers ( EEEA-Net-C2, OFA-595, DenseNet161, Visformer-S, and  AutoFormerV2-T) on AUC over ChestX-ray14~\cite{Wang2017ChestXRay8} dataset.}
\label{tab:3}
\begin{tabular}{lccccc}
\hline
  &  \multicolumn{5}{c}{\textbf{Teachers}}  \\  
\textbf{Pathologies} & \textbf{EEEA-Net-C2 \cite{termritthikun2021eeea}} & \textbf{OFA-595 \cite{cai2020once}} & \textbf{DenseNet161 \cite{huang2017densely}} & \textbf{Visformer-S \cite{chen2021visformer}} & \textbf{AutoFormerV2-T \cite{chen2021autoformerv2}} \\ \hline
Atelectasis & 78.10 & 79.46 & 80.86 & \textbf{80.91} & 79.72 \\
Cardiomegaly & 89.37 & 90.40 & \textbf{91.71} & 90.60 & 90.86 \\
Consolidation & 78.67 & 79.41 & \textbf{79.95} & 79.44 & 79.81 \\
Edema & 89.50 & 89.58 & \textbf{89.63} & 88.56 & 88.71 \\
Effusion & 86.72 & 87.75 & \textbf{88.21} & 87.50 & 87.80 \\
Emphysema & 90.00 & 91.41 & \textbf{92.30} & 92.20 & 91.51 \\
Fibrosis & 75.90 & 78.02 & \textbf{78.91} & 77.79 & 77.49 \\
Hernia & 85.41 & 91.37 & \textbf{94.48} & 90.22 & 90.01 \\
Infiltration & 70.17 & 71.19 & 71.03 & \textbf{71.30} & 70.76 \\
Mass & 80.55 & 82.24 & \textbf{85.24} & 84.89 & 82.75 \\
Nodule & 72.96 & 74.51 & \textbf{78.57} & 77.57 & 73.70 \\
Pleural\_Thickening & 77.67 & 79.42 & \textbf{80.19} & 79.85 & 77.15 \\
Pneumonia & 75.22 & 75.69 & \textbf{76.17} & 73.78 & 75.76 \\
Pneumothorax & 83.63 & 85.29 & \textbf{86.33} & 86.00 & 85.93 \\ \hline
Mean AUC & 80.99 & 82.55 & \textbf{83.83} & 82.90 & 82.28 \\ \hline
\end{tabular} 

\centering
\vspace{0.5cm}
\caption{Label-wise performance of EEEA-Net-C2 students in AUC on ChestX-ray14~\cite{Wang2017ChestXRay8} dataset. Second column shows the baseline, while third column shows label-wise AUC of EEEA-Net-C2 student trained by various teachers (EEEA-Net-C2, OFA-595, DenseNet161, Visformer-S, and AutoFormerV2-T).}
\label{tab:4}
\begin{tabular}{l|c|ccccc}
\hline
  & \textbf{Baseline} & \multicolumn{5}{c}{\textbf{Teachers}}  \\  
\textbf{Pathologies} & \textbf{EEEA-Net-C2 \cite{termritthikun2021eeea}} & \textbf{EEEA-Net-C2 \cite{termritthikun2021eeea}} & \textbf{OFA-595 \cite{cai2020once}} & \textbf{DenseNet161 \cite{huang2017densely}} & \textbf{Visformer-S \cite{chen2021visformer}} & \textbf{AutoFormerV2-T \cite{chen2021autoformerv2}} \\ \hline
Atelectasis & 78.10 & 79.41 & 80.08 & 80.50 & \textbf{80.63} & 80.06 \\
Cardiomegaly & 89.37 & 90.60 & 91.01 & \textbf{92.04} & 90.84 & 91.65 \\
Consolidation & 78.67 & 79.53 & 79.78 & 79.98 & 80.00 & \textbf{80.29} \\
Edema & 89.50 & 90.06 & \textbf{90.18} & 89.94 & 89.67 & 89.81 \\
Effusion & 86.72 & 87.62 & 87.88 & 87.96 & 87.84 & \textbf{88.28} \\
Emphysema & 90.00 & 91.47 & 92.40 & 91.88 & \textbf{92.46} & 92.29 \\
Fibrosis & 75.90 & 76.75 & 78.27 & 79.04 & \textbf{79.28} & 78.20 \\
Hernia & 85.41 & 90.95 & 93.72 & \textbf{94.40} & 90.29 & 92.28 \\
Infiltration & 70.17 & 70.92 & \textbf{71.67} & 71.54 & 71.45 & 71.11 \\
Mass & 80.55 & 82.56 & 83.58 & \textbf{84.57} & 84.55 & 83.79 \\
Nodule & 72.96 & 75.04 & 76.02 & \textbf{77.40} & 76.34 & 76.18 \\
Pleural\_Thickening & 77.67 & 78.64 & \textbf{80.44} & 80.08 & 79.36 & 78.57 \\
Pneumonia & 75.22 & 75.82 & 75.67 & \textbf{77.01} & 75.24 & 76.78 \\
Pneumothorax & 83.63 & 85.94 & 86.45 & 86.15 & 85.46 & \textbf{86.57} \\ \hline
Mean AUC & 80.99 & 82.52 & 83.37 & \textbf{83.75} & 83.10 & 83.28 \\ \hline
\end{tabular}
\end{table*}

\subsection{Implementation} 

Both the teacher and student models are built using native PyTorch. We used the official implementations with default settings unless otherwise specified. All experiments are performed on four NVIDIA V100 GPUs. AdamW optimizer is used for optimization where the input image resolution is maintained at 224$\times$224 with a batch size of 512. The teacher model is, first, trained for ten epochs. Then, the teacher model is employed to train the student model for another ten epochs. Each of the datasets, \ie, ChestX-ray14, CheXpert, and PadChest, is divided with the ratio of 80\% and 20\% into the training and validation sets for training and validation, respectively. 

All utilized teacher networks in the proposed method are pre-trained models trained on the ImageNet dataset (RGB format). However, the datasets (ChestX-ray14, CheXpert, and PadChest) utilized by the proposed method contain grayscale images. To avoid altering the architecture of each pre-trained model, we devised a strategy to transform all greyscale (224$\times$224$\times$1) images of three datasets to RGB (224$\times$224$\times$3) format to match the pre-trained model format.

\subsection{Impact of CNN- and Transformer-based Teachers}
\label{Section:Experiment:ImpactofCNNTransTeacher}

In this section, we confirm that student performance can be improved by both the CNN and Transformer teachers. At first, Table~\ref{tab:1} shows the performance of the following neural networks, \ie,  EEEA-Net-C2~\cite{termritthikun2021eeea}, OFA-595~\cite{cai2020once}, DenseNet161~\cite{huang2017densely}, Visformer-S\cite{chen2021visformer}, and  AutoFormerV2-T~\cite{chen2021autoformerv2}. Different neural networks yield the best performance on different evaluation metrics and datasets. On ChestX-ray14, DenseNet161 performs the best on AUC and F1; OFA-595 and AutoFormerV2-T provide the best Accuracy and F1 scores. On ChestXpert, OFA-595 performs the best on Accuracy and F1, while Visformer-S yields the best AUC and F1. On PadChest, Visformer-S provides the best AUC, while DenseNet161 provides the best Accuracy and F1. Notice that  EEEA-Net-C2 offers the moderate performance with the lowest number of parameters and FLOPS among other neural networks. Thus,  we use EEEA-Net-C2 offers as the student in the next study.

We study the improved performance of the students in Table~\ref{tab:2} using OFA-595, DenseNet161, Visformer-S, and  AutoFormerV2-T as the teacher. From Table~\ref{tab:2}, the student performance is improved by KD from both CNN and Transformer teachers, especially, DenseNet161, Visformer-S, and AutoFormerV2-T. On ChestX-ray14, EEEA-Net-C2 is improved by DenseNet161 as the teacher on AUC by 2.7\%; and it is improved by AutoFormerV2-T as the teacher on Accuracy and F1 by 0.5\%  and 0.4\%, respectively.   On ChestXpert, EEEA-Net-C2 is improved by Visformer-S as the teacher on AUC by 1.8\%; and it is improved by DenseNet161 as the teacher on Accuracy and F1 by 1.0\% and 0.2\%, respectively. On PadChest, EEEA-Net-C2 is improved by Visformer-S across all the three metrics, \ie, 3.0\% on AUC, 0.2\% on accuracy, and 0.3\% on F1; it is also improved by DenseNet161 on accuracy and F1 by 0.2\%  and 0.3\%, respectively, and by Visformer-S for 0.2\% on accuracy.

\begin{figure*}[t]
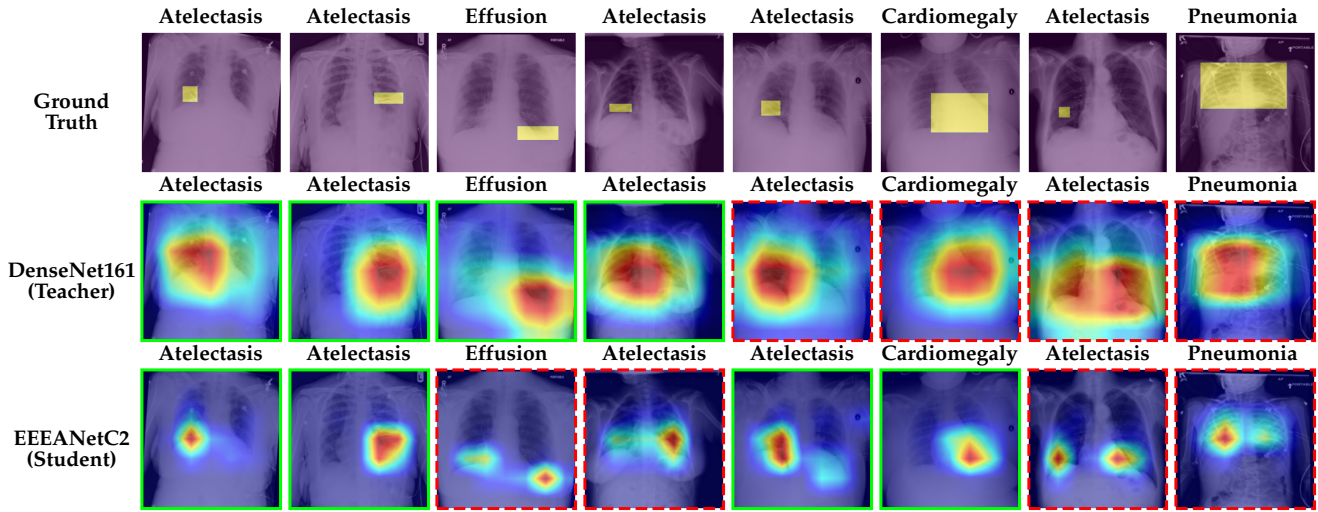
      
    \renewrobustcmd{\bfseries}{\fontseries{b}\selectfont}    
    {\fontsize{8}{8}\selectfont   
        {  \setlength\tabcolsep{0.5pt}      \renewcommand{\arraystretch}{1.1} 
            \begin{tabular}{ P{1.7cm} cccccccc     }  %   
                \DTLforeach{fig2}{\meth=meth,\colA=colA,\colB=colB,\colC=colC,\colD=colD,\colE=colE,\colF=colF,\colG=colG,\colH=colH}{   
                    \ifthenelse{\value{DTLrowi}=1}{\tabularnewline  }{\tabularnewline}   
                    \ifthenelse{\value{DTLrowi}=1 \OR \value{DTLrowi}=3  \OR \value{DTLrowi}=5  }{ 
                        \bfseries \meth     &  \bfseries \colA  &  \bfseries  \colB  & \bfseries \colC & \bfseries \colD & \bfseries \colE & \bfseries \colF & \bfseries \colG & \bfseries \colH    }{    }
                    \ifthenelse{ \value{DTLrowi}=2  \OR \value{DTLrowi}=4  \OR \value{DTLrowi}=6  }{   
                        \bfseries \meth     &
                        \includegraphics [align=c, width=0.106\textwidth, trim={0.05cm 0.05cm 0.05cm 0.05cm},clip] {\colA}  &   
                        \includegraphics [align=c,width=0.106\textwidth, trim={0.05cm 0.05cm 0.05cm 0.05cm},clip] {\colB}  & 
                        \includegraphics [align=c,width=0.106\textwidth, trim={0.05cm 0.05cm 0.05cm 0.05cm},clip] {\colC} &
                        \includegraphics [align=c,width=0.106\textwidth, trim={0.05cm 0.05cm 0.05cm 0.05cm},clip] {\colD} &
                        \includegraphics [align=c,width=0.106\textwidth, trim={0.05cm 0.05cm 0.05cm 0.05cm},clip] {\colE} &
                        \includegraphics [align=c,width=0.106\textwidth, trim={0.05cm 0.05cm 0.05cm 0.05cm},clip] {\colF} &
                        \includegraphics [align=c,width=0.106\textwidth, trim={0.05cm 0.05cm 0.05cm 0.05cm},clip] {\colG}&
                        \includegraphics [align=c,width=0.106\textwidth, trim={0.05cm 0.05cm 0.05cm 0.05cm},clip] {\colH}   }{    } 
                }
            \end{tabular} 
        }     
    }
    \caption{Heatmap of teacher (DenseNet161) and student (EEEA-Net-C2). \textit{Solid green} and \textit{dashed red} frames denote   \textit{correct} and \textit{incorrect} classification respectively. Four pair-wise combinations of correct and incorrect examples are shown between teacher and student network.}   
    \label{fig:2}    
\end{figure*}

\subsection{Label-wise Improvement}
\label{Section:Experiment:Labelwise}

In this section, we present the label-wise performance improvement by KD on ChestX-ray14~\cite{Wang2017ChestXRay8}. At first, Table~\ref{tab:3} provides the label-wise, baseline performance of the following neural networks, \ie,  EEEA-Net-C2, OFA-595, DenseNet161, Visformer-S, and AutoFormerV2-T. DenseNet161 offers the highest AUC in most cases, \ie, it offers the highest AUC on 12 out of 14 pathologies. Visformer-S offers the second best performance where it achieves the highest AUC on 2 out of 14 pathologies. 

Then, the improved label-wise performance by KD is provided in Table~\ref{tab:4} where EEEA-Net-C2 is the student. Generally, DenseNet161 is still the best teacher in improving the mean AUC from the student baseline's by 2.38\%. Notice that the EEEA-Net-C2 students whose teachers are OFA-595, Visformer-S, and AutoFormerV2-T, offer higher mean AUC than their own teachers. In addition, these students provide the higher performance in classifying some pathologies. For example, the EEEA-Net-C2 student offers higher AUC than its OFA-595 teacher by 13 out of 14 pathologies. The EEEA-Net-C2 student of Visformer-S teacher offers higher AUC than its teacher by 9 out of 14 pathologies. The EEEA-Net-C2 student of AutoFormerV2-T teacher offers higher AUC than its teacher for all 14 pathologies.

\subsection{Explainable AI for Knowledge Distillation}
\label{Section:Experiment:ExplainableAI}

To interpret the transferring of knowledge from teachers to students, we study the visualization of the heatmap on ChestX-ray14~\cite{Wang2017ChestXRay8}. We use the ground-truth bounding box from ChestX-ray14 to denote the area corresponding to the identified pathologies. We employed Grad-CAM \cite{selvaraju2017gradcam} to provide the heatmap of the identified pathologies.

\begin{figure*}[!t]
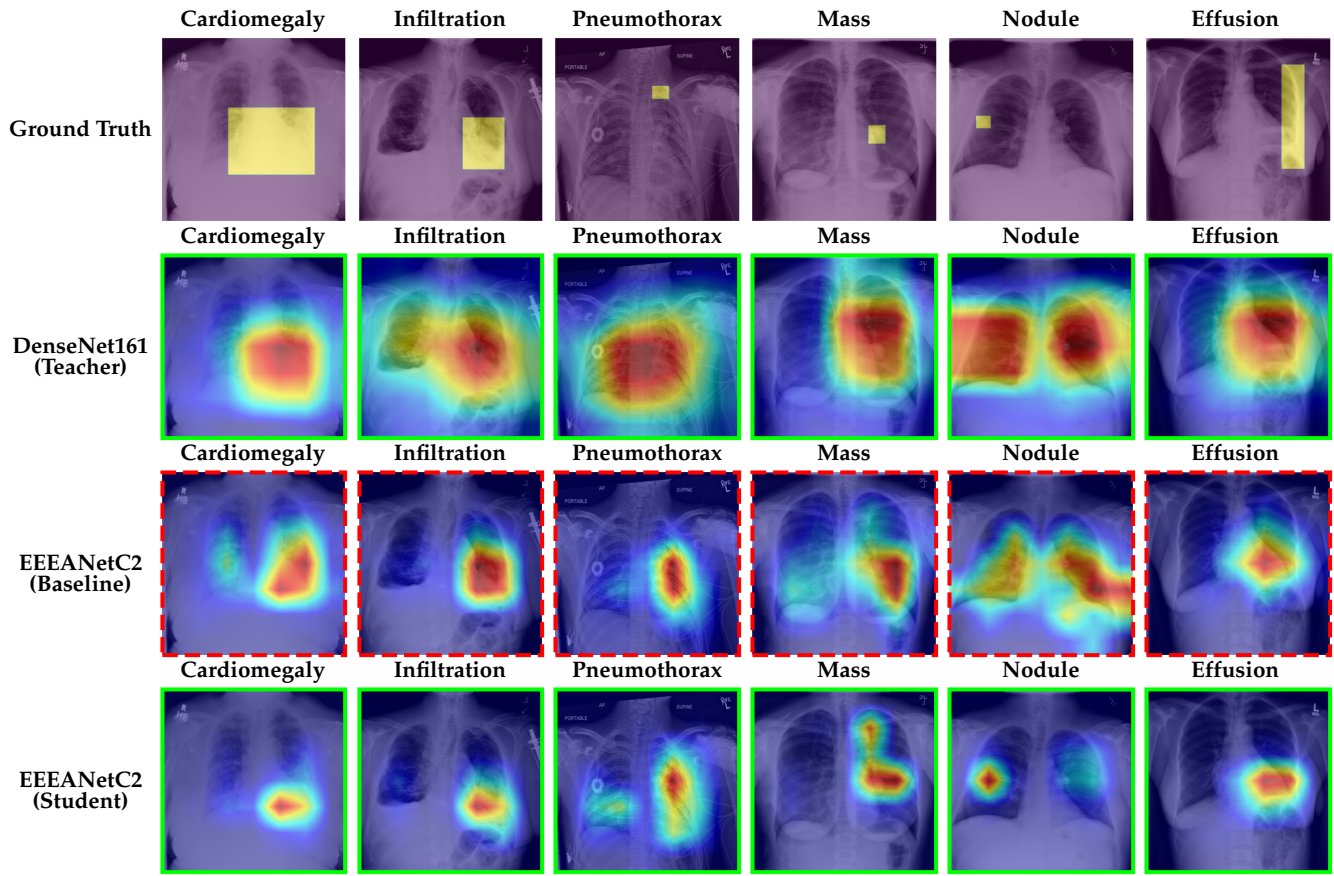
      
    \renewrobustcmd{\bfseries}{\fontseries{b}\selectfont}    
    {\fontsize{8.5}{8}\selectfont   
        {  \setlength\tabcolsep{1pt}    \renewcommand{\arraystretch}{1.2} 
            \begin{tabular}{ P{1.9cm} ccccccc     }  %   
                \DTLforeach{fig3}{\meth=meth,\colA=colA,\colB=colB,\colC=colC,\colD=colD,\colE=colE,\colF=colF}{   
                    \ifthenelse{\value{DTLrowi}=1}{\tabularnewline  }{\tabularnewline}   
                    \ifthenelse{\value{DTLrowi}=1 \OR \value{DTLrowi}=3  \OR \value{DTLrowi}=5 \OR \value{DTLrowi}=7}{ 
                        \bfseries \meth     &  \bfseries \colA  &  \bfseries  \colB  & \bfseries \colC & \bfseries \colD & \bfseries \colE & \bfseries \colF}{    }
                    \ifthenelse{ \value{DTLrowi}=2 \OR \value{DTLrowi}=4  \OR \value{DTLrowi}=6 \OR \value{DTLrowi}=8   }{   
                        \bfseries \meth     &
                        \includegraphics [align=c, width=0.14\textwidth, trim={0.05cm 0.05cm 0.05cm 0.05cm},clip] {\colA}  &    
                        \includegraphics [align=c,width=0.14\textwidth, trim={0.05cm 0.05cm 0.05cm 0.05cm},clip] {\colB}  & 
                        \includegraphics [align=c,width=0.14\textwidth, trim={0.05cm 0.05cm 0.05cm 0.05cm},clip] {\colC}  & 
                        \includegraphics [align=c,width=0.14\textwidth, trim={0.05cm 0.05cm 0.05cm 0.05cm},clip] {\colD}  & 
                        \includegraphics [align=c,width=0.14\textwidth, trim={0.05cm 0.05cm 0.05cm 0.05cm},clip] {\colE}  & 
                        \includegraphics [align=c,width=0.14\textwidth, trim={0.05cm 0.05cm 0.05cm 0.05cm},clip] {\colF}  }{  
                        } 
                }
            \end{tabular} 
        }     
    }
    \caption{The visualization demonstrating the improved performance of the student, denoted with the \textit{solid green} and \textit{dashed red} frame for the \textit{correct} and \textit{incorrect} classification. From top to bottom are the ground-truth bounding box (representing radiologist perspective), DenseNet161 (teacher), EEEA-Net-C2 (baseline), and EEEA-Net-C2 (student).}   
    \label{fig:3}    
\end{figure*}

Fig.~\ref{fig:2} provides the heatmap visualization from the student and teacher networks, denoted with the \textit{green} and \textit{red} frames for the \textit{correct} and \textit{incorrect} classification. Four combination cases of classification by teacher and student networks are analyzed, with each case including two examples of samples classified as Atelectasis, Effusion, Cardiomegaly, and Pneumonia. Fig.~\ref{fig:2} demonstrates the effectiveness of KD for various correct/incorrect classification combinations. DenseNet161 is used as the teacher network, while EEEA-Net-C2 is used as student network, where EEEA-Net-C2 employs KD by utilizing DenseNet161 as the teacher network. From the visualizations it can be observed that, the teacher's correct classification corresponds to the large heatmap that covers the ground-truth bounding boxes of Atelectasis and Effusion.
Meanwhile, the student's correct classification corresponds to a much smaller heatmap that intersects with the ground-truth bounding boxes such as Atelectasis and Cardiomegaly. On the other hand, the incorrect classification of teacher on Atelectasis, Cardiomegaly, and Pneumonia is associated to the heatmap whose peak deviates from the ground-truth bounding box. Meanwhile, the incorrectness of the student network corresponds to the heatmap with additional peaks that deviate from the locations of the ground-truth bounding box. 

\begin{table}[t]
\centering
\vspace{0.5cm}
\caption{Evaluation of the proposed method (Knowledge Distillation) of Teacher-Student combinations for ChextX-ray14~\cite{Wang2017ChestXRay8}, which shows the quantitative knowledge transferred from teacher to the student.}
\label{tab:6}
\begin{tabular}{ccc}
\hline
\textbf{DenseNet161 (teacher)} & \textbf{EEEA-Net-C2 (student)} & \textbf{Percentage(\%)} \\ \hline
\multicolumn{1}{c|}{Correct}   & \multicolumn{1}{c|}{Correct}   & 50.7                    \\
\multicolumn{1}{c|}{Correct}   & \multicolumn{1}{c|}{Incorrect} & 4                       \\
\multicolumn{1}{c|}{Incorrect} & \multicolumn{1}{c|}{Correct}   & 6.5                     \\
\multicolumn{1}{c|}{Incorrect} & \multicolumn{1}{c|}{Incorrect} & 38.8                    \\ \hline
\end{tabular}
\end{table}

Table~\ref{tab:6} shows the evaluation of teacher-student pairings when KD is utilized for the dataset ChestX-ray14. DenseNet161 is utilized as the teacher network, whereas EEEA-Net-C2 is used as the student network. ChestX-ray14 has 112,120 CXR images, where 984 CXR images have bounding box available as a segmentation mask. For a fair assessment, a refinement is done by only considering the clean CXR images out of these 984 images. The selection criteria consider only the samples which are identical in both segmentation $(BBox \_List \_2017)$ and original ChestX-ray14 $(Data \_Entry \_2017)$ datasets. The refined dataset contains 201 clean CXR images, where these images have identical labels in both segmentation and ChextX-ray14 datasets. In addition, the classification results for all clean 201 CXR images are collected in a new CSV file along with the class of interest for both the teacher (DenseNet161) and student networks (EEEA-Net-C2). Table~\ref{tab:6} is created by utilizing 201 clean CXR images, label of interest, and the output probabilities of both the teacher and student models. Finally, teacher-student pairing combinations are considered to quantify the effectiveness of KD approach by providing teacher (DenseNet-161) and student (EEEA-Net-C2) models evaluation for multi-label CXR image classification in the form of correct/incorrect classification labels. This provides further insight into the quantitative knowledge transferred from teacher to student model by categorizing the number of correct/incorrect classification labels. From the results, it can be observed that when teacher output classification is correct, 92.68\% time it transfers the correct knowledge, whereas 7.3\% wrong knowledge is transferred to student network.
Furthermore, Fig.~\ref{fig:3} provides the visualization demonstrating the improved performance of the student by the transferred knowledge from the teacher with respect to the student baseline: first row represents radiologist perspective in the form of bounding box, while second and fourth rows demonstrate the teacher and student model's response against the radiologist findings in the form of heatmaps respectively. We selected 6 samples identified with Cardiomegaly, Infiltration, Pneumothorax, Mass, Nodule, and Effusion. With KD, the peak of the heatmap of the EEEA-Net-C2 student is shifted from its baseline and becomes closer to either the teacher's area or the ground-truth bounding box. On Cardiomegaly, Infiltration, and Effusion, the peak of heatmap moves towards the teacher's area. On Pneumothorax, Mass, and Nodule, the peak of the heatmap moves closer to the ground-truth bounding box. This confirms the impact of the supervision by both the teacher's and the ground truth label.          

\begin{table}[t]
\centering
\caption{On-device performance comparison between the student EEEA-Net-C2 versus the  teachers: OFA-595, DenseNet161, Visformer-S, and AutoFormerV2-T, for CXR image classification.  }
\label{tab:5}
\begin{threeparttable}
{
\begin{tabular}{lc|P{0.8cm}P{0.5cm}P{0.5cm}P{1cm}} 
\hline 
&       &   \multicolumn{4}{c}{\textbf{Latency}}\\  \cline{3-6}
&    \textbf{Size}  &  \textbf{SDM845}  &  \textbf{Intel}  &  \textbf{Apple}  &  \textbf{NVIDIA} \\
\textbf{Model} &  \textbf{(MB)} &  \textbf{(ms)}  &  \textbf{(ms)}  &  \textbf{(ms)}  &  \textbf{(ms)}  \\ \hline
EEEA-Net-C2 \cite{termritthikun2021eeea} & \textbf{18.3} & \textbf{107} & \textbf{88.4} & \textbf{97.0} & \textbf{5.2} \\
OFA-595 \cite{cai2020once} & 29.4 & 134 & 237.1 & 124.8 & 6.5 \\
DenseNet161 \cite{huang2017densely} & 102 & 663 & 894.8 & 612.6 & 28.8 \\
Visformer-S \cite{chen2021visformer} & 150 & 361 & 283.1 & 333.4 & 15.2 \\
AutoFormerV2-T \cite{chen2021autoformerv2} & 107 & 348 & 245.4 & 322.3 & 9.7 \\ \hline
\end{tabular} 
}
\begin{tablenotes} 
    \item  SDM845 represents Qualcomm Snapdragon 845; Intel represents Intel Core i7-6700HQ CPU; Apple represents Apple M1 Pro CPU; NVIDIA represents NVIDIA GTX 980 Ti GPU. 
\end{tablenotes} 
\end{threeparttable}
\end{table}

\subsection{On-device Performance}
\label{Section:Experiment:OnDevicePerformance}
Table~\ref{tab:5} provides the comparison of the on-device performance, \ie, model size, and latency, between students and teachers. We deploy both the student and teachers on different hardware, \ie,  Qualcomm Snapdragon 845, Intel Core i7-6700HQ CPU, Apple represents Apple M1 Pro CPU, and NVIDIA represents NVIDIA GTX 980 Ti GPU. The student EEEA-Net-C2 had the smallest model size and latency. Among all the teachers, Visformer-S requires the largest memory size. Meanwhile, DenseNet161 has the highest latency across different hardware, \ie, about 6, 10, 6, and 5 times of the student's inference runtime, on Qualcomm Snapdragon 845, Intel Core i7-6700HQ CPU, Apple M1 Pro, and NVIDIA GTX 980 Ti, despite the fact that DenseNet161 generally offers the highest AUC in multi-label classifications. This confirms the advantage of our KD strategy that offers the better alternative, \ie, the EEEA-Net-C2, which is computationally efficient and offers high accuracy.

\section{Conclusion}
\label{Section:Conclusion}
Diagnosis of chest-related disorders using CXR images is a difficult task that demands particular attention from the scientific community given the disease's increasing prevalence. We investigated KD for CXR image analysis and discovered that we were able to successfully transfer knowledge from a teacher model to a student model without modifying the original student model. Additionally, we analyze XAI by providing a visual explanation of our proposed KD technique. The experimental results show that when DenseNet161 is used as a teacher to train the EEEA-Net-C2 model with KD, an AUC of 83.7\%, 87.1\%, and 88.7\% is seen across three datasets, including ChestX-ray14, CheXpert, and PadChest. Furthermore, it can be observed that when teacher output classification is correct, 92.68\% time it transfers correct knowledge, whereas 7.3\% wrong knowledge is transferred to student network. Moreover, our model EEEA-Net-C2 is a compact model that is compatible with a wide range of constrained hardware platforms.
 
One limitation of the proposed work is the information regarding ground truth localization (i.e., bounding box) is not yet readily available; meanwhile, manual annotation from an expert radiologist is an expensive and time-consuming task. In the future, one could increase the dataset size for improved model performance using weakly supervised learning to obtain ground truth localization. 

The proposed method can be applied to semi-supervised learning (pseudo labels) due to the limitation of a small dataset. XAI used in the proposed method can be extended to image segmentation, and object detection within medical imaging.

\section*{Acknowledgement}
This work is funded in part by Australian Government Research Training Program (RTP) Scholarship.

\ifCLASSOPTIONcaptionsoff
  \newpage
\fi

\bibliographystyle{IEEEtran}
\bibliography{mybibfile}

\begin{IEEEbiography}[{\includegraphics[width=1in,height=1.25in,clip,keepaspectratio]{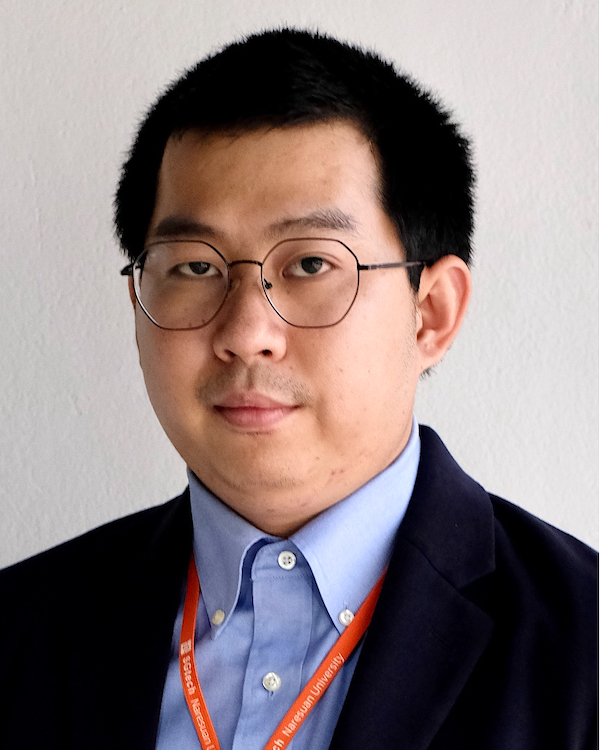}}]
{Chakkrit~Termritthikun} (M'22) is a Lecturer and Assistant Director for Digital Innovation at the School of Renewable Energy and Smart Grid Technology, Naresuan University, Thailand. He received the B.Eng., M.Eng., and Ph.D. degrees in computer engineering from Naresuan University in 2013, 2017, and 2021, respectively. From 2018 to 2020, he was a Visiting Research Student at the University of South Australia, Adelaide, Australia. His current research interests include deep learning, neural architecture search, on-device artificial intelligence, and blockchain.
\end{IEEEbiography}

\begin{IEEEbiography}[{\includegraphics[width=1in,height=1.25in,clip,keepaspectratio]{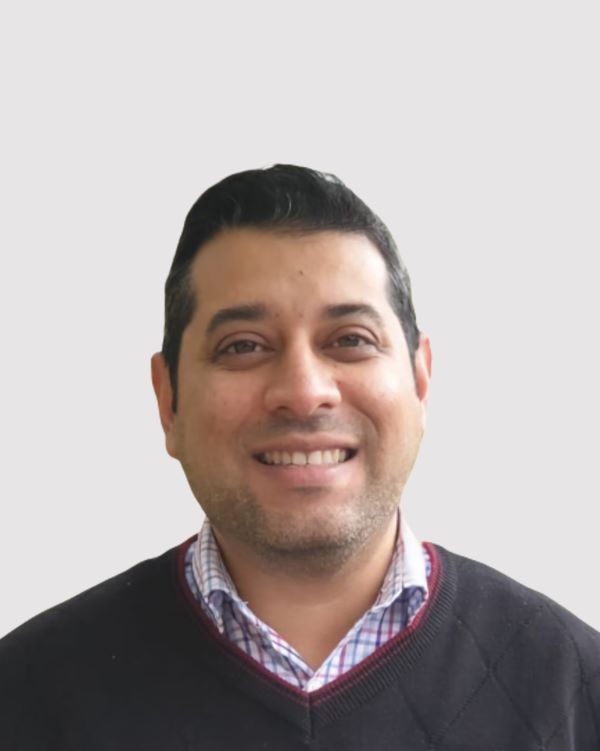}}]
{Ayaz Umer} received the B.Eng. degree in information technology from the University of Engineering and Technology Taxila, Pakistan in 2005, and the M.Eng. degree in telecommunications from the University of South Australia in 2008, and he is currently working toward the Ph.D. degree in IT at the same university. He has worked as a Lecturer with the Department of electrical engineering, COMSATS University, Attock Campus, Pakistan. He has worked at various positions including teaching at undergraduate level and industrial research projects (Research Themes Investment Scheme) with UniSA. His current research interests include computer vision, deep learning, and visual attention.
\end{IEEEbiography}

\begin{IEEEbiography}[{\includegraphics[width=1in,height=1.25in,clip,keepaspectratio]{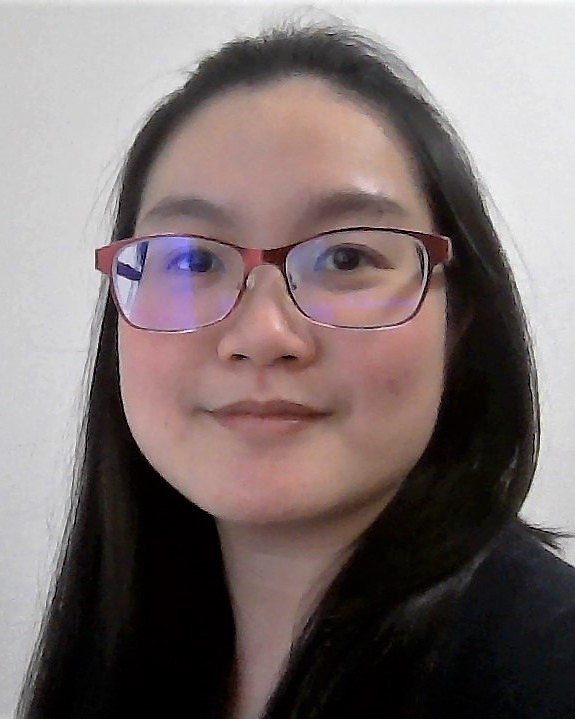}}]
  {Suwichaya Suwanwimolkul} (M'22) is an independent researcher. She received a Ph.D. degree in computer science from The University of Adelaide, Australia, in 2018. Her research interests include visual localization, compressive sensing, and machine learning. 
\end{IEEEbiography}

\begin{IEEEbiography}[{\includegraphics[width=1in,height=1.25in,clip,keepaspectratio]{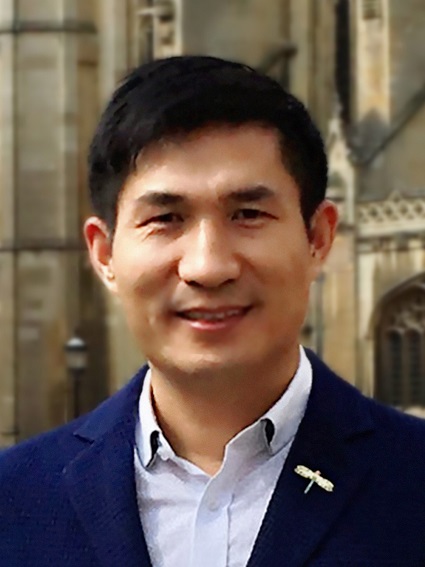}}]
  {Feng Xia} (M'07-SM'12) received the BSc and PhD degrees from Zhejiang University, Hangzhou, China. He is a Professor in School of Computing Technologies, RMIT University, Australia. Dr. Xia has published 2 books and over 300 scientific papers in international journals and conferences (such as IEEE TAI, TKDE, TNNLS, TC, TMC, TPDS, TBD, TCSS, TNSE, TETCI, TETC, THMS, TVT, TITS, TASE, ACM TKDD, TIST, TWEB, TOMM, WWW, AAAI, SIGIR, WSDM, CIKM, JCDL, EMNLP, and INFOCOM). His research interests include data science, artificial intelligence, graph learning, and systems engineering. He is a Senior Member of IEEE and ACM, and an ACM Distinguished Speaker.
\end{IEEEbiography}

\begin{IEEEbiography}[{\includegraphics[width=1in,height=1.25in,clip,keepaspectratio]{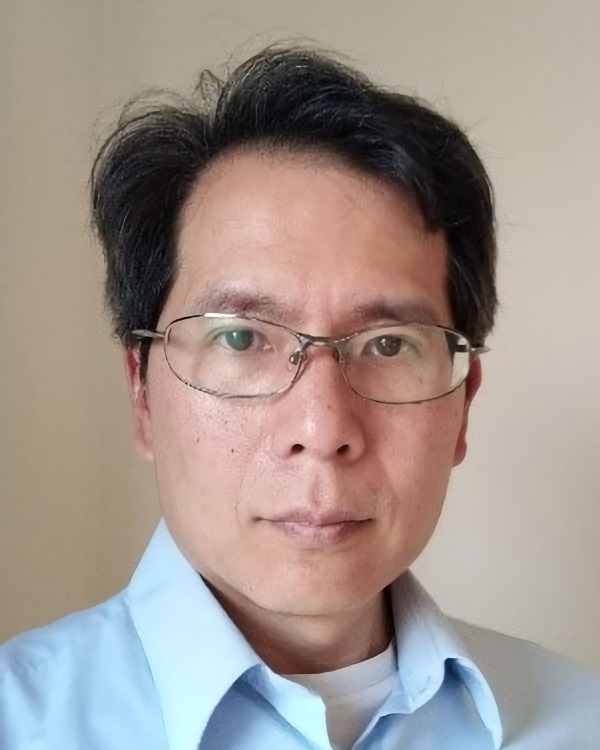}}]
  {Ivan~Lee} received the B.Eng., M.Com., MER, and Ph.D. degrees from The University of Sydney. He worked at Cisco Systems, Remotek Corporation, and Ryerson University. He is currently an Associate Professor at the University of South Australia, and a REDI Fellow. His research interests include intelligent sensors, multimedia systems, and data analytics.
\end{IEEEbiography}

\end{document}